# Real-Time Super-Resolution System of 4K-Video Based on Deep Learning

## (Invited Paper)

Yanpeng Cao[1], Chengcheng Wang[1], Changjun Song[1], Yongming Tang[1], He Li[2]
[1]*Joint International Research Laboratory of Information Display and Visualization, Southeast University,* Nanjing, CN
[2]*Department of Engineering, University of Cambridge,* Cambridge, UK
{caoyp, zebra, songcj, tym}@seu.edu.cn, he.li@ieee.org

*Abstract*—Video super-resolution (VSR) technology excels in reconstructing low-quality video, avoiding unpleasant blur effect caused by interpolation-based algorithms. However, vast computation complexity and memory occupation hampers the edge of deplorability and the runtime inference in real-life applications, especially for large-scale VSR task. This paper explores the possibility of real-time VSR system and designs an efficient and generic VSR network, termed EGVSR. The proposed EGVSR is based on spatio-temporal adversarial learning for temporal coherence. In order to pursue faster VSR processing ability up to 4K resolution, this paper tries to choose lightweight network structure and efficient upsampling method to reduce the computation required by EGVSR network under the guarantee of high visual quality. Besides, we implement the batch normalization computation fusion, convolutional acceleration algorithm and other neural network acceleration techniques on the actual hardware platform to optimize the inference process of EGVSR network. Finally, our EGVSR achieves the real-time processing capacity of 4K@29.61FPS. Compared with TecoGAN, the most advanced VSR network at present, we achieve 85.04% reduction of computation density and 7.92× performance speedups. In terms of visual quality, the proposed EGVSR tops the list of most metrics (such as LPIPS, tOF, tLP, *etc.*) on the public test dataset Vid4 and surpasses other state-of-the-art methods in overall performance score. The source code of this project can be found on https://github.com/Thmen/EGVSR.

## I. INTRODUCTION

Video super-resolution (VSR) is developed from image super-resolution, and it is one of the hot topics in the field of computer vision. VSR technology reconstructs degraded video, restores the definition of video, and improves the subjective visual quality. VSR is of great significance for improving the quality of early low-resolution video resources. At present, high-resolution such as 4K or even 8K display technology is relatively mature, however, the mainstream video sources are still dominated by low-resolution such as 1080P or 720P, which limits the quality of video system from the source end. In the near future, 4K and even higher resolution will surely replace Full HD (FHD) as the mainstream format. Therefore, there is an urgent real-life demand for efficient and lightweight VSR technology to upgrade a mass of low-resolution (LR) videos to high-resolution (HR) ones.

The research object of VSR technology is the image sequence of video resources. The image sequence is composed of a series of static images and forms into continuous frames. Since some objects in the video move at a fast speed and appear as a motion blur effect in a single image, there will be sub-pixel displacements between target frames and its adjacent frames. Therefore, it is crucially important for VSR systems to align the adjacent frames using effective motion compensation algorithms. This is a difficult and challenging problem in the field of current VSR research. In addition, super-resolution technology still has the following formidable challenges and urgent research directions:

- **Large scale and unknown corruption,** still lack of effective algorithms.
- **Lightweight and real-time architecture**, where deep VSR models are still difficult to deploy on hardware.

Deep Learning algorithms are considered to be excel at solving many unsupervised problems, and essential to solve the aforementioned challenges. This paper explores the solutions of large-scale VSR and pursues the goal of 4K high-resolution in VSR system.

For large-scale VSR, challenges in the computational complexity and memory consumption impede the real-time and low latency performance of video processing. Although advanced deep models have achieved high quality on VSR, these models are still difficult to be deployed in practical applications due to the huge amount of parameters and calculations. In order to solve this problem, we need to design a lightweight VSR model, or refine the existing VSR model with fewer parameters and sparse structures.

Generally, in the field of VSR, main research direction lies in the pursuit of video quality, while few focus on fast and real-time VSR methods. Real-time VSR requires to consider both of quality and speed. In this paper, we propose a VSR network that can handle large-scale and high-performance, and investigate hardware-friendly accelerating architectures for VSR network inference, thereby allowing real-time processing without the sacrifice of VSR quality. The contributions of this paper are summarised as follows:

1. We present a lightweight and efficient VSR network to improve the performance of VSR quality and running speed.
2. We investigate various network acceleration strategies tailored for large-scale VSR system to meet the requirements of real-time inference.
3. We propose an unified method to quantify different metrics of VSR quality for efficient automated evaluation across vast test samples.

## II. RELATED WORK

### A. Deep Learning Based Video Super Resolution

From the perspective of technical route, super-resolution (SR) technology can be summarized into three categories: interpolation based SR, super-resolution reconstruction based

SR, and learning based SR [1]. In the last few years, interests in deep learning (DL) based SR algorithms research have risen rapidly. It is difficult for traditional algorithms to make breakthroughs for higher performance, while DL-based SR algorithms have achieved significant improvements of SR quality [2]. In addition, compared to single-image SR, video SR problems provide more available information from multiple frames, with both spatial dependence of intra-frame and temporal dependence of inter-frame. Therefore, the existing work mainly focuses on how to make an efficient use of spatio-temporal correlation, which refers to explicit motion compensation (MC) and recursive feedback mechanism to fuse additional image information from multi-frames.

In terms of MC based VSR methods, Liao *et al.* [3] used multiple optical flow methods to generate HR candidate objects and integrated them into CNNs. VSRnet [3] estimated the optical flow through the Druleas algorithm, SOFVSR [4] reconstructed the coarse-to-fine optical flow through the OFRnet network. Both used multiple continuous frames as the input of CNNs to predict HR frames. Besides, some methods tried to learn MC directly. VESPCN [5] used a trainable spatial transformer to learn MC between adjacent frames, and input multiple frames into a spatio-temporal network ESPCN [6] for end-to-end prediction. BRCN [7] proposed a bi-directional framework that using CNN, RNN, and conditional Generative Adversarial Network (GAN) for model spatial, temporal, and spatio-temporal dependence, respectively. FRVSR [8] and TecoGAN [9] used the previous HR predicted frames to reconstruct the subsequent HR frames in a circular manner through two DNNs. Another trend started to use recursive method to capture spatio-temporal correlations without the need for explicit MC. Specifically, DUF [10] used an end-to-end deep CNN to generate dynamic upsampling filters and residual images to avoid explicit MC processing. EDVR [11] used the enhanced deformable convolutions and RBPN [12] utilized a recurrent encoder-decoder module to improve the fusion of multi-frame information.

### B. Efficient and Real-time VSR Network

Following the design principle of CNN networks, "the deeper, the better", VSR networks have been developing towards a larger and wider network architecture. However, large-scale networks bring huge computation, making it difficult to be implemented on present-constrained hardware platforms and deploy practical VSR networks in real-time. Recently, many research studies have investigated optimization and acceleration methods of VSR network. For example, Chao *et al.* redesigned and optimized network structure in order to accelerate the previous SRCNN model [13], therefore, the network complexity of FSRCNN is much lower than that of SRCNN. FAST [14] used compression algorithm to extract a compact description of the structure and pixel correlation, and accelerated the most advanced SR algorithm by 15 times with a minimum performance loss (only -0.2 dB). The VSRnet proposed by Kappeler *et al.* used an adaptive MC architecture to deal with motion blur problems, and the processing time of each frame only needs 0.24s on GPU device [15].

Furthermore, interests in FPGA-based high-performance and parallel computing have grown. In the early work [16, 17], researchers first implemented large-scale VSR tasks on FPGA, *i.e.* 2Kto8K@60Hz 4× video upscale and 4Kto8K@60Hz 2× upscale, however, they still used the non-DL traditional interpolation-based algorithm. The energy-efficient DCNNs devised by Chang *et al.* optimized the deconvolutional layer, and proposed the FPGA-based CNN accelerator to generate UHD video efficiently [18]. Under the same occupation of hardware resources, the throughput of the DCNN accelerator is 108 times faster than a traditional implementation. Yongwoo *et al.* proposed a hardware-friendly VSR network based on FPGA facilitated by quantisation and network compression [19].

### III. OUR METHODS

#### A. Efficient and Generic VSR System

The generative and discriminative modules of GANs can play games with each other during the training process to produce better perceptual quality than traditional generative models. Therefore, GANs are widely used in the SR field. We rely on the powerful ability of deep feature learning of GAN models to deal with large-scale and unknown degradation challenges in VSR tasks. In addition, we refer to the design of the TecoGAN [9] method and introduce the spatio-temporal adversarial structure to help the discriminator understand and learn the distribution of spatio-temporal information, which avoids instability effect in temporal domain encountered by traditional GANs.

Moreover, to meet the requirements of super-resolving large-scale video up to 4K-resolution, we follow the design principles of efficient CNN model to build a more generic and high-quality video super-resolution network, termed EGVSR (efficient and generic video super-resolution). To allow processing 4K video in real-time, we recall the practical guidelines of an efficient CNN architecture [20] and build an lightweight network structure for EGVSR. The generator part is divided into FNet module and SRNet module for optical flow estimation and video frame super-resolution, respectively. Figure 1 shows the framework of EGVSR's generator part and the data flow during inference stage.

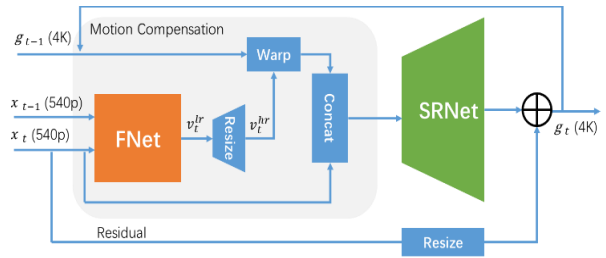

Fig. 1. Overview of EGVSR network.

The structure of FNet refers to the encoder-decoder architecture in RNN to estimate the dense optical flow and provide motion compensation information for adjacent frame alignment operation (Warp). The encoding part uses three encoder units, each of which is composed of {Conv2d→LeakyReLU→Conv2d→LeakyReLU→MaxPool2}, and the decoding part uses three decoder units formed by {Conv2d→LeakyReLU→Conv2d→LeakyReLU→BilineraUp×2}. The design of SRNet module needs to take into account both network capacity and inference speed. We remark that multiple network layers are used to ensure the VSR quality, while the network complexity should be controlled for real-time video processing ability.

Herein, we refer to the structure of ResNet [21] network and adopt lightweight residual block (ResBlock) to build SRNet. The structure of ResBlock is {(Conv2d→ReLU→Conv2d)+Res}. Considering the balance between quality and speed, we use 10 ResBlock to build SRNet, and use lightweight sub-pixel convolutional layer in the end of EGVSR network as the output upsampling method, with the structure of {PixelShuffle×4→ReLU→Conv2d}.

As for the other modules in our EGVSR, we keep the same setup to the previous work [9] for fair comparison, considering the balance between the inference speed of the EGVSR network and the quality of the VSR. Moreover, a lightweight network is introduced without affecting the quality of the VSR. The design principle is to simplify the EGVSR network as much as possible, and uses the various neural network acceleration techniques mentioned below.

*B. Batch Normalization Fusion*

In order to ensure real-time processing capability of our EGVSR system, further optimizations are made in EGVSR system without sacrificing the quality of VSR. Batch Normalization (BN) technology is most commonly used in the field of deep learning to improve the generalization of the network and prevent the side effect of over-fitting. The mathematical formula for the calculation of BN processing can be briefly described as Eq. (1). It can be seen that the calculation of BN is quite complicated, and the mean ($\mu$) and variance ($\sigma^2$) value of a batch of samples need to be counted first. The FNet module in our EGVSR network also makes extensive use of the BN layer. We need to optimize it to improve the speed of network training and inferencing.

$$\hat{x}_i = \frac{\gamma x_i}{\sqrt{\sigma^2}} + (\beta - \frac{\gamma \mu}{\sqrt{\sigma^2 + \grave{o}}})$$
$$\mu = \frac{1}{n}\sum_{i=1}^{n} x_i, \sigma^2 = \frac{1}{n}\sum_{i=1}^{n}(x_i - \mu)^2 \quad (1)$$

First of all, we transform the BN calculation into matrix form, as expressed in (2):

$$\begin{pmatrix} \hat{F}_{1,i,j} \\ \hat{F}_{2,i,j} \\ \vdots \\ \hat{F}_{C,i,j} \end{pmatrix} = \begin{pmatrix} \frac{\gamma_1}{\sqrt{\hat{\sigma}_1^2+\epsilon}} & 0 & \cdots & 0 \\ 0 & \frac{\gamma_2}{\sqrt{\hat{\sigma}_2^2+\epsilon}} & & \vdots \\ \vdots & & \ddots & 0 \\ 0 & \cdots & 0 & \frac{\gamma_C}{\sqrt{\hat{\sigma}_C^2+\epsilon}} \end{pmatrix} \begin{pmatrix} F_{1,i,j} \\ F_{2,i,j} \\ \vdots \\ F_{C,i,j} \end{pmatrix} + \begin{pmatrix} \beta_1 - \frac{\gamma_1 \hat{\mu}_1}{\sqrt{\hat{\sigma}_1^2+\epsilon}} \\ \beta_2 - \frac{\gamma_2 \hat{\mu}_2}{\sqrt{\hat{\sigma}_2^2+\epsilon}} \\ \vdots \\ \beta_C - \frac{\gamma_C \hat{\mu}_C}{\sqrt{\hat{\sigma}_C^2+\epsilon}} \end{pmatrix} \quad (2)$$

We can see that the transformed BN layer is similar to the formation of the 1×1 convolution $f(\vec{x}) = W * \vec{x} + b$, then we can utilize the 1×1 convolutional layer to realize and replace the layer of BN. Finally, we can fuse the 1×1 convolutional layer with the previous convolutional layer, so that we can eliminate the need of calculating BN. The optimization of BN fusion will provide a speed improvement of about 5%. The overall transformation process is shown in Figure 2.

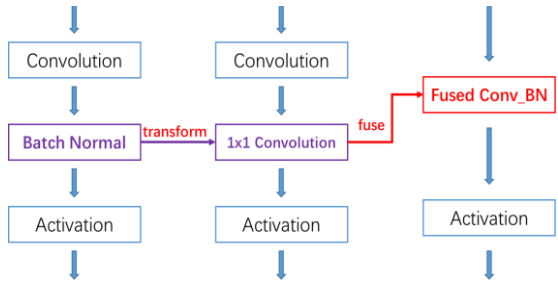

Fig. 2. Batch Normalization fusion processing flow.

*C. Efficient Upsampling Method*

Upsampling layer is one of the most important portions in SR network, which can be roughly divided into two categories according to different technique routes: traditional interpolation-based upsampling methods and learning-based upsampling methods. All interpolation upsampling methods have an obvious defect, which will cause edge blur to the image, while various learning-based upsampling methods, such as Deconvolution, Sub-pixel Convolution, and Resize Convolution, have powerful feature learning capability, and show their talents in VSR networks.

In order to compare the above three intra-network upsampling methods to select the best solution, we used the control variate method to evaluate the efficiency of these upsampling layers in actual SR networks. Specifically, ESPCN [6] network is used as the backbone of SR network. We only changed the upsampling layer while maintaining the other network structures and configurations, and trained multiple groups of SRNet with three different upsampling methods, *i.e.* **A.** Resize convolution (using bilinear interpolation); **B.** Deconvolution; **C.** Sub-pixel convolution. Table I shows the detailed network settings.

TABLE I. THE NETWORK SETTING OF THREE CONTROL SRNETS

| Structure | Layer | Output Shape | Param# |
|---|---|---|---|
| Backbone | 1-Conv2d+Tanh | [1,64,800,800] | 1,664 |
| | 2-Conv2d+Tanh | [1,32,800,800] | 18,464 |
| | 3-Conv2d+Tanh | [1,32,800,800] | 9,248 |
| Upsample-**A** | 4-Interpolation | [1,32,2400,2400] | 0 |
| | 5-Conv2d | [1,1,2400,2400] | 33 |
| Upsample-**B** | 5-ConvTranspose2d | [1,1,2400,2400] | 801 |
| Upsample-**C** | 4-Conv2d | [1,9,800,800] | 297 |
| | 5-PixelShuffle | [1,1,2400,2400] | 0 |

Table II records the performance metrics of different SRNets. It can be seen that the sub-pixel convolution has the best quality performance in both the training and testing stages, except for PSNR metric in testing stage, which is slightly lower than that of deconvolution (-0.02dB). Besides, we test the average running time of different SRNets for 3× super-resolving single test image with the size of 800×800 under the same testing environment. Although the SRNet with resize convolution has the least weight parameters, the processing time bottleneck lies in both CPU and GPU platforms, due to the high computational complexity of interpolation. Sub-pixel convolution performs better than method **A** and **B**, which is 1.77 times faster (CPU) and 1.58 times faster (GPU) than that of method **A**.

TABLE II. EXPERIMENTAL COMPARISON RESULTS OF VARIOUS UPSAMPLING METHODS

| Up-sample Method | Total Param# | Train | | Test | | CPU time (ms) | GPU time (ms) |
|---|---|---|---|---|---|---|---|
| | | Loss | PSNR | PSNR | SSIM | | |
| **A** | 29,409 | 0.0055 | 22.61 | 25.45 | 0.72 | 415.8 | 9.860 |
| **B** | 30,177 | 0.0048 | 23.20 | 26.52 | 0.76 | 253.4 | 8.203 |
| **C** | 29,673 | 0.0047 | 23.28 | 26.50 | 0.77 | 234.9 | 6.234 |

*D. Convolutional Computation Acceleration*

In order to further improve the inference speed of EGVSR network, we explore the core of computation in the neural network. From the perspective of actual engineering deployment, it points out that convolutional computation is the key to CNNs, accounting for more than 90% of the total computation, which consumes most of the calculation time, therefore, it is necessary to improve the computational efficiency of convolution. We should design an efficient convolutional algorithm suitable for hardware deployment.

According to basic calculation process of the traditional naïve convolution (ConV), a large number of loop structures (6 loops) are used, and the computation efficiency is quite low. In terms of two-dimensional convolutional layer with a 3×3 kernel, we need to traverse from top left to bottom right on the input feature map based on the traditional sliding window method to obtain the output feature map, as shown in Fig. 3.

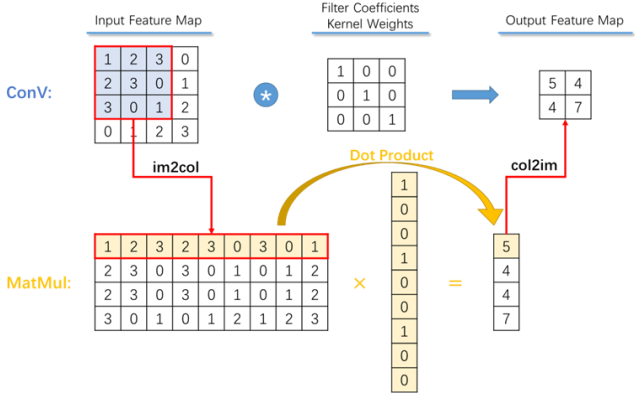

Fig. 3. Use matrix multiplication to accelerate convolutional computation.

We consider using the matrix multiplication (MatMul) algorithm to improve it, following the process in the second line of Figure 3. First, input data of each activation zone is extracted according to the size of convolutional filter, and converted from original 2D matrix with 3×3 size into single row vector. When the length of sliding step is one, there are four activation zones in total, thus four 1D row vectors can be obtained. All vector constitutes a 2D matrix with a size of 4×9. We call this conversion as im2col (*i.e.* feature map to column vector or image to column), and this optimization method was first proposed by Jia [22]. Similarly, the 2D convolutional filter with the size of 3×3 is straightened directly and transformed into the 1D column vector with the size of 9×1. Such a conversion does not consume computation, and it is only a rearrangement of the memory in reading order. There will be duplicated data elements in the converted matrix, which will increase the memory usage.

We find that MatMul computation of two transformed matrices identify with the results of a convolutional computation, and no additional calculation is required. The desired output feature results can be obtained through the inverse col2im conversion. The convolution operation in CNN is essentially a multiple dimensional correlation computation. In our actual hardware implementation, the method mentioned above is adopted to convert convolutional computation into matrix multiplication, which saves inference time by memory space to boost higher computational efficiency.

We also aim to accelerate our proposed EGVSR network on FPGAs using convolution accelerators. We recall our previous work WinoConv [23], a FPGA-based convolution accelerator, and analyse the feasibility of EGVSR's edge deployment on FPGA, where Winograd algorithm [24] is used to reduce the complexity of convolutional computation, decreased from $O(n^3)$ to $O(n^{2.376})$.

## IV. EXPERIMENTS AND DISCUSSION

### A. Evaluation of Image Quality

Firstly, we evaluated and compared the actual super-resolution performance of image quality on the standard testing dataset VID4 with previous VSR networks under different technical routes, including: 1).VESPCN [5] and SOFVSR [4] networks based on non-GAN method with MC; 2). DUF [10] network based on non-GAN method without explicit MC; and 3). our EGVSR network based on GAN method and optical flow based MC.

In order to facilitate comparison, we captured the actual image results from different VSR methods and focus more on the detail textural features. Figure 4 exhibits the reconstruction results produced by various VSR networks on VID4 dataset, and the group of detail images on the right side represents the image results from LR (low-resolution), VESPCN, SOFVSR, DUF, EGVSR and GT (ground-truth) respectively. From the subjective results, EGVSR's results are the closest to the GT images and achieve higher image detail reconstruction quality. VESPCN and SOFVSR networks performed relatively fuzzy in the overall picture and seriously lost most image edge details. EGVSR against the DUF network that currently has state-of-the-art performance of image quality in VSR field.

For a more objective assessment of VSR image quality, we used three most common metrics: PSNR, SSIM and LPIPS. The specific experimental results are shown in Table III. The objective testing results are consistent with the previous subjective results, and it can be seen that DUF and EGVSR seem equally matched in three metrics. Generally, DUF performs slightly better in PSNR and SSIM metrics, while our EGVSR performs better in LPIPS. Regarding to the evaluation of image super-resolving quality, Blau and Michaeli have proved that the measurement using PSNR or SSIM metric to assess the human visual perception quality has an inherent distortion [25]. DL-based feature mapping metric LPIPS can capture more high-level image semantic structures, and the LPIPS metric is close to the subjective evaluation of human eyes. Therefore, LPIPS is more accurate than the first two metrics, and our EGVSR has a significant performance improvement of 48.15% compared with DUF in LPIPS, according to the average results on the VID4 dataset.

TABLE III. OBJECTIVE EVALUATION RESULTS OF IMAGE QUALIT ON VID4 TEST DATASET

| Sequence Name | Metric | VESPCN | SOFVSR | DUF | **Ours** |
|---|---|---|---|---|---|
| Calendar | PSNR↑ | 14.67 | 18.39 | 23.59 | 23.60 |
| | SSIM↑ | 0.19 | 0.50 | 0.80 | 0.80 |
| | LPIPS↓ | 0.57 | 0.41 | 0.33 | 0.17 |
| City | PSNR↑ | 19.38 | 22.03 | 27.63 | 27.31 |
| | SSIM↑ | 0.14 | 0.69 | 0.79 | 0.79 |
| | LPIPS↓ | 0.48 | 0.21 | 0.27 | 0.16 |
| Foliage | PSNR↑ | 16.22 | 22.96 | 26.15 | 24.79 |
| | SSIM↑ | 0.09 | 0.46 | 0.77 | 0.73 |
| | LPIPS↓ | 0.54 | 0.36 | 0.35 | 0.14 |
| Walk | PSNR↑ | 15.28 | 20.91 | 29.90 | 27.84 |
| | SSIM↑ | 0.32 | 0.45 | 0.91 | 0.86 |
| | LPIPS↓ | 0.34 | 0.44 | 0.14 | 0.09 |
| Average | PSNR↑ | 16.20 | 21.02 | 26.82 | 25.88 |
| | SSIM↑ | 0.19 | 0.53 | 0.82 | 0.80 |
| | LPIPS↓ | 0.48 | 0.36 | 0.27 | 0.14 |

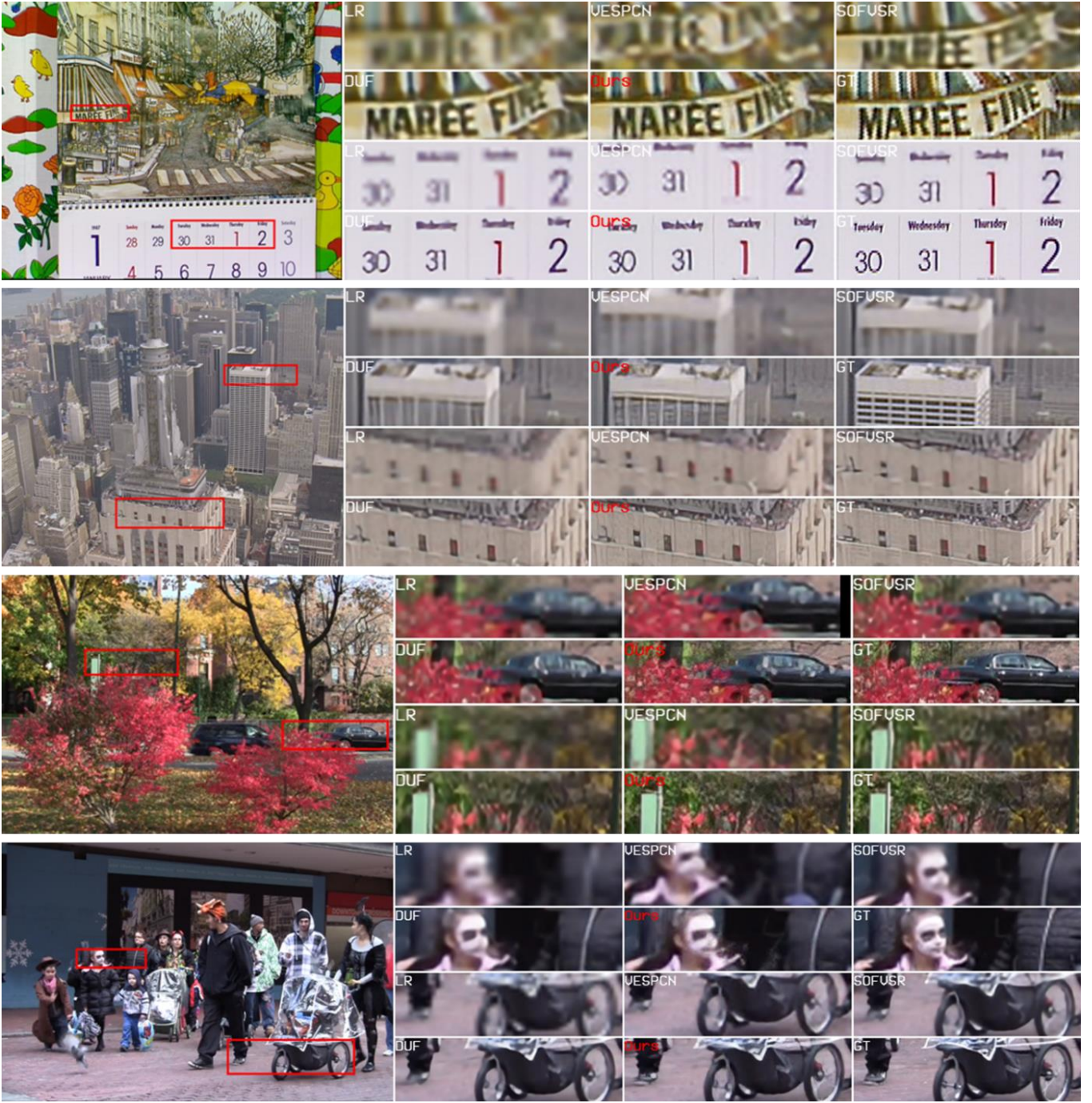

Fig. 4. Subjective comparison results from various VSR methods (Testing on Vid4 dataset, video sequence name: Calendar/City/Foliage/Walk).

## B. Evaluation of Video Quality and Temporal Coherence

In this section, we will investigate the objective evaluation of video quality for our VSR system. In order to acquire the qualitative comparison result of temporal consistency, we introduced two temporal metrics, tOF and tLP, used in previous work [9]. tOF measures the pixel-wise difference of motions estimated from sequences, and tLP measures perceptual changes over time using deep feature map:

$$\begin{aligned} \text{tOF} &= \left\| OF(b_{t-1}, b_t) - OF(g_{t-1}, g_t) \right\|_1 \\ \text{tLP} &= \left\| LP(b_{t-1}, b_t) - LP(g_{t-1}, g_t) \right\|_1 \end{aligned} \quad (3)$$

Pixel differences and perceptual variations are critical to quantifying actual temporal coherence, therefore, tOF and tLP will measure the difference between the VSR results and the corresponding GT reference ones. The smaller the score is, the closer it is to the true result, which providing a more pleasant and fluent subjective perceptual experience. In addition to the VESPCN and SOFVSR networks (DUF is excluded, mainly because it is not based on explicit MC), two latest VSR networks, FRVSR [8] and TecoGAN [9], are used. We conducted testing experiment on three datasets, VID4, TOS3 and GVT72. The specific experiment results are shown in Figure 5 and Figure 6. The results of the temporal metrics show that the spatio-temporal adversarial model has better performance than the traditional model. EGVSR network can recover more spatial details with a satisfied temporal coherent, meeting the subjective perception of human eyes.

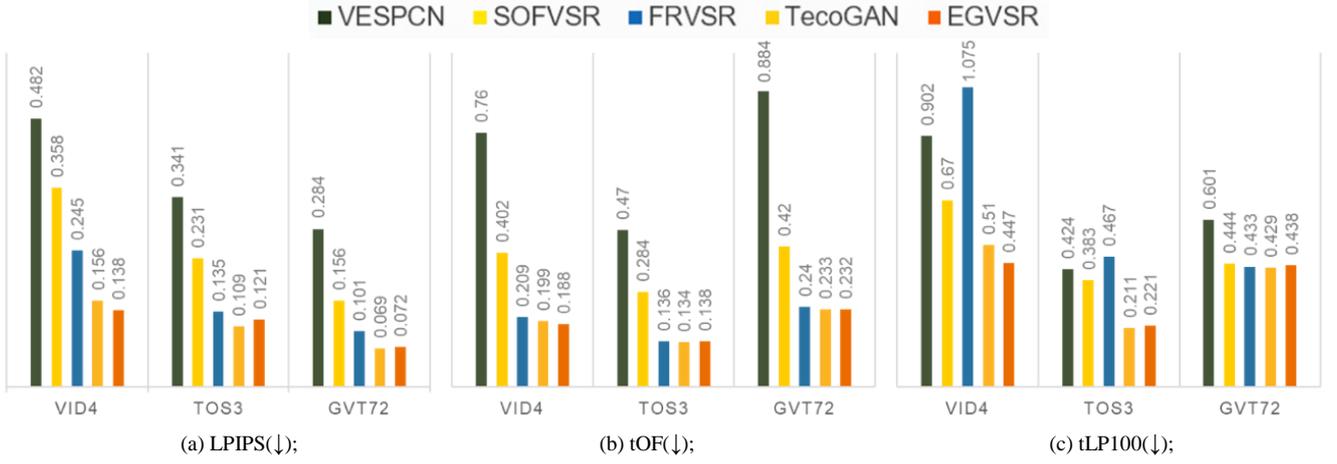

Fig. 5. Averaged VSR metric evaluations for three dataset with the following metrics：LPIPS, tOF, tLP.

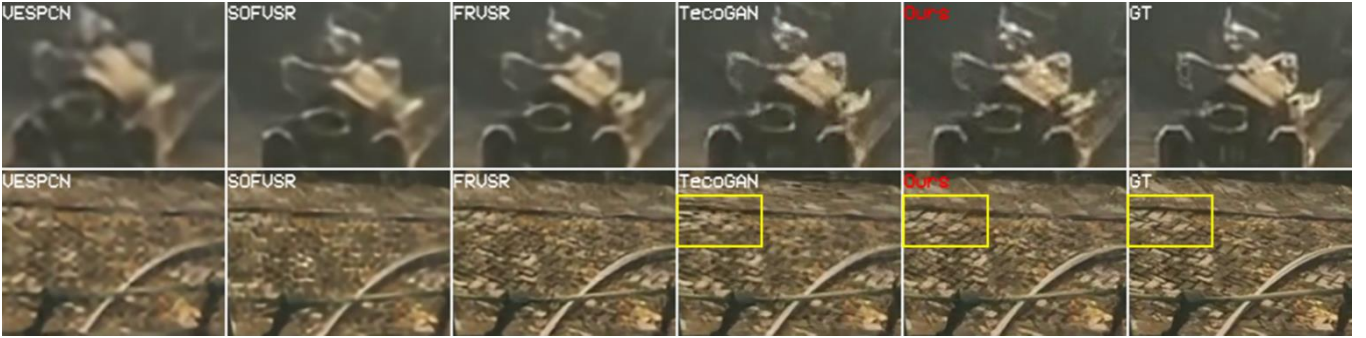

Fig. 6. VSR comparisons for detial views of captured image ("Bridge" video sample in TOS3 dataset) in order to compare to previous work.

TecoGAN model stands out in all temporal performance test of TOS3 dataset. There is still a slight gap between our EGVSR and TecoGAN model, where a reduction performance of -4.74%~-11.01% is shown in evaluation result. However, it is difficult to distinguish their difference from the aspect of subjective perception, as shown in Figure 6. Besides, EGVSR is even slightly better in some representation details, such as the reconstruction of brick texture details marked in the yellow box of the "Bridge" sample in Figure 6. Images generated by EGVSR are closer to the GT ones. EGVSR has an advantage over TecoGAN in some respects, or even overall exceeds TecoGAN on VID4, maintaining a performance advantage of +5.53% to +12.35%. The performance of our EGVSR network in temporal domain is significantly better than that of the previous methods and is comparable to that of TecoGAN, the SOTA VSR model by far.

### C. Runtime Speed on GPU

This section will test the running frame rate of different VSR models during inference. The experimental results are shown in Table IV. The 2nd line lists the parameters of each VSR networks, line 3 counts the statistics of corresponding computation cost, and the last two lines show the average FPS that can be achieved during 4× VSR in different resolutions. The results show that, the total computation cost required by our EGVSR is only 29.57% of VESPCN, 12.63% of SOFVSR, and 14.96% of FRVSR and TecoGAN. In terms of using CPU only, we achieve the increase of speed-up by 8.25× to 9.05× compared to TecoGAN. As for using GPU for acceleration, we realize the EGVSR system in real-time at a speed of 720p/70.0FPS, 1080p/66.9FPS, 4K/29.6FPS, which has 2.25×, 4.45×, and 7.92× performance speed-up compared to TecoGAN. Due to our lightweight design of VSR network and various optimization strategies, the efficiency of EGVSR on CPU/GPU hardware platform is improved greatly. In contrast, other DL-based VSR methods have limited runtime efficiency when dealing with large-scale VSR tasks such as 1080p and 4K resolution, which cannot meet the runtime ability (above 25FPS).

### D. Overall Performance

Although the above experimental discussion provides the test result for evaluating the visual quality and running speed of our VSR system, these test experiments are compared on their own dimension independently. An unified quantitative metric is essential for efficient automated evaluation across a large number of test samples. In this section, we consider the balance between visual quality and running speed of VSR network for generating high-resolution video. Therefore, we propose a novel and unified VSR visual quality assessment metric to quantify LPIPS in spatial domain and tOF and tLP in temporal domain. Specifically, Eq. (4) is used to normalize the value of all metrics of each network in different datasets.

$$M_{nor} = (M - M_{\min})/(M_{\max} - M_{\min}) \quad (4)$$

The weighted sum method is used to quantify different metrics, and finally the comprehensive visual quality score of VSR network is calculated by

$$Score = 1 - \sum_{i=1}^{n}(\lambda_i M_i^{nor}) \quad (5)$$

where, the value of the score ranges from 0 to 1, a higher score indicating that the VSR system achieves a better visual quality.

Figure 7 depicts the comprehensive performance of video quality score and network running speed of various VSR methods. In addition to VESPCN, SOFVSR, DUF, FRVSR,

TABLE IV. THE RUNTIME SPEED OF DIFFERENT VSR NETWORKS ON CPU AND GPU

| Performance | Source | Target | VESPCN | SOFVSR | FRVSR | TecoGAN | **Ours: EGSVR** | Speed-up vs. TecoGAN |
|---|---|---|---|---|---|---|---|---|
| Parameters(M) | -- | -- | 0.879 | 1.640 | 2.589 | 2.589 | 2.681 | |
| FLOPs(G) | 320×180 | 720p | 96.56 | 226.12 | 190.81 | 190.81 | 28.55 | -- |
| | 480×270 | 1080p | 221.08 | 508.78 | 429.30 | 429.30 | 64.06 | -- |
| | 960×540 | 4K | 886.47 | 2035.11 | 1718.65 | 1718.65 | 257.01 | -- |
| FPS(CPU) | 320×180 | 720p | 3.053 | 1.039 | 1.152 | 1.150 | 9.487 | 8.25× |
| | 480×270 | 1080p | 1.201 | 0.443 | 0.485 | 0.485 | 4.389 | 9.05× |
| | 960×540 | 4K | 0.289 | 0.106 | 0.112 | 0.112 | 0.958 | 8.55× |
| FPS(GPU) | 320×180 | 720p | 48.48 | 13.31 | 31.16 | 31.15 | 70.04 | 2.25× |
| | 480×270 | 1080p | 24.76 | 5.993 | 15.10 | 15.05 | 66.90 | 4.45× |
| | 960×540 | 4K | 6.78 | 1.734 | 3.76 | 3.74 | 29.61 | 7.92× |

TABLE V. SYNTHESIS RESULTS ON FPGA

| Input Size | Method 2019 [26] LUT-based Direct Convolution | | | Method 2017 [27] DSP-based Direct Convolution | | | | **Ours: WinoConv** LUT-based Winograd Convolution | | | Max FLOPs (T) |
|---|---|---|---|---|---|---|---|---|---|---|---|
| | FF | LUT | Latency | DSP | FF | LUT | Latency | FF | LUT | Latency | |
| 4×4 | 191 | 493 | 39 | 2 | 383 | 658 | 11 | 343 | 827 | 6 | 2.839 |
| 5×5 | 243 | 635 | 85 | 3 | 681 | 2055 | 22 | 1138 | 2682 | 10 | 0.821 |
| 6×6 | 253 | 666 | 148 | 4 | 853 | 2888 | 31 | 1794 | 4242 | 12 | 0.623 |
| 7×7 | 233 | 654 | 229 | 5 | 1150 | 3834 | 48 | 5111 | 10214 | 16 | 0.264 |
| 8×8 | 233 | 515 | 328 | 6 | 1239 | 4862 | 60 | 8055 | 16499 | 17 | 0.201 |

TecoGAN and EGVSR mentioned above, we also obtained the specific performance of EDVR [11] and RBPN [12] from the public data. In terms of running speed, the average running of various VSR networks on the GPU for 4× video super-resolution with target resolution of 4K is tested. As shown in Figure 7, the closer to the upper right corner, the better visual performance and faster running speed can VSR network achieve. The color and size of the bubble represents the computational complexity and parameter number of network, respectively. In summary, the overall visual quality of EGVSR network is at the advanced level, second only to TecoGAN network (lower 0.011/1.14%), while it is the only VSR network that is capable of processing 4K video in real-time (29.61FPS).

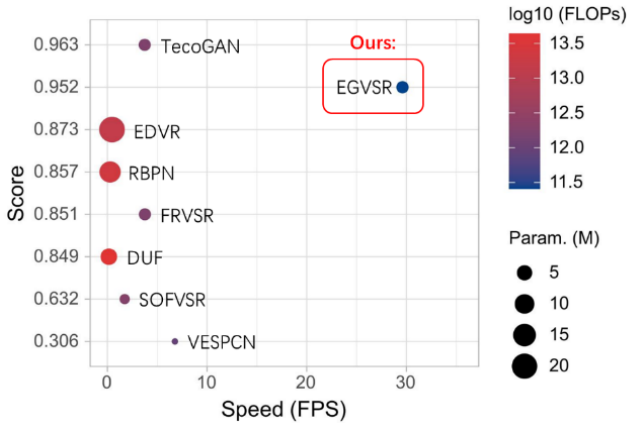

Fig. 7. Overall performance comparison of various VSR networks.

### E. FPGA Deployment Estimation

We have deployed the prototype design of the convolution accelerator, WinoConv mentioned above, on a Xilinx KC705 development platform. And, in this section, we tested and evaluated the WinoConv convolution accelerator on KC705 under 300MHz. We compared our WinoConv with previous work: LUT-based convolution method [26] and DSP-based convolution method [27]. Table V shows the hardware synthesis results of different methods to achieve 3×3 convolution.

Experimental results show that, WinoConv has the lowest computational latency and has great advantages in terms of convolutional computation speed. The delay of LUT-based direct convolution method is unacceptable among three methods. Compared to DSP-based convolution method, our method can reduce the latency at least 1.83×, and yields more speed-up gains with larger convolution size. Besides, we have calculated the max FLOPs by the following formulas:

$$FLOPs_{wino} = C_i \cdot H_i \cdot W_i \cdot K^2 \cdot C_o \quad (6)$$

$$MAX_FLOPs = \frac{\#LUT_{total}}{\#LUT_{wino}} \times FLOPs_{wino} \times \frac{Frequency}{Latency_{wino}} \quad (7)$$

The last column of Table V indicates the maximal FLOPs provided by different WinoConv accelerators. Combined with the computation cost required by the EGVSR network given in Table IV, the implementation of the whole EGVSR network on FPGA edge deployment could realize the runtime speed of 720p/99.44FPS, 1080p/44.32FPS, 4K/11.05FPS in the way of theoretical estimation. We remark that implementing the entire VSR system on FPGAs would meet the demands of edge and low-energy computing, as a task in the future.

## V. CONCLUSIONS

In this paper, we have conducted an in-depth study in the VSR field to address the 4K-resolution VSR tasks and efficient VSR processing in real-time. Using various optimization strategies, the proposed EGVSR method reduces the computation load to the lowest requirement, under the premise of high visual quality of VSR, and realizes a real-time 4K VSR implementation on hardware platforms. The balance between quality and speed performance is improved effectively. Even though we have designed the accelerator for convolutional computation on FPGAs, while it is considerable to deploy the whole system on FPGA platform to further achieve the possibility of edge inference for VSR systems.